\documentclass{article}
\usepackage{spconf,amsmath,graphicx}
\usepackage{multirow}

\title{F-PABEE: Flexible-patience-based Early Exiting for Single-label and Multi-label text Classification Tasks}
\name{Xiangxiang Gao$^1$, Wei Zhu$^2$\sthanks{Wei Zhu contributes equally with Xiangxiang Gao, and he is the corresponding author. Email: wzhu@stu.ecnu.edu.cn. }, Jiasheng Gao$^3$, Congrui Yin$^4$}
\address{$^1$ Shanghai Jiaotong University, China \\
 $^2$ East China Normal University, China \\ $^3$ Shenzhen University, China \\ $^4$ Nanchang University, China}

\begin{document}
%
\maketitle
\begin{abstract}

Computational complexity and overthinking problems have become the bottlenecks for pre-training language models (PLMs) with millions or even trillions of parameters. A Flexible-Patience-Based Early Exiting method (F-PABEE) has been proposed to alleviate the problems mentioned above for single-label classification (SLC) and multi-label classification (MLC) tasks. F-PABEE makes predictions at the classifier and will exit early if predicted distributions of cross-layer are consecutively similar. It is more flexible than the previous state-of-the-art (SOTA) early exiting method PABEE because it can simultaneously adjust the similarity score thresholds and the patience parameters. Extensive experiments show that: (1) F-PABEE makes a better speedup-accuracy balance than existing early exiting strategies on both SLC and MLC tasks. (2) F-PABEE achieves faster inference and better performances on different PLMs such as BERT and ALBERT. (3) F-PABEE-JSKD performs best for F-PABEE with different similarity measures.

\end{abstract}
\section{Introduction}

Fine-tuning PLMs has become the de-facto paradigm in natural language processing~\cite{DBLP:journals/corr/abs-2106-04554}, due to the amazing performance gains on a wide range of natural language processing tasks \cite{Zhu2020MVPBERTRV,Zhu2019PANLPAM,Zuo2022ContinuallyDR,Zhu2023ACFAC,Guo2021GlobalAD}. Despite SOTA performances, BERT~\cite{DBLP:journals/corr/abs-1810-04805} and its variants~\cite{DBLP:journals/corr/abs-1909-11942, DBLP:journals/corr/abs-1906-08237, DBLP:journals/corr/abs-1907-11692,zhu-2021-mvp} still face significant application challenges: cumbersome computation and overthinking problems due to huge parameters and deep models. Early exiting attracts much attention as an input-adaptive method to speed up inference~\cite{earlyexitingreview}. Early exiting installs a classifier at each transformer layer to evaluate the predictions and will exit when meeting the criterion. Three different early exiting strategies exist: (1) The confidence-based strategy evaluates the predictions based on specific confidence measurements. (2) The learned-based strategy learns a criterion for early exiting. (3) The patience-based strategy exits when consecutive classifiers make the exact predictions. Among them, the patience-based strategy PABEE~\cite{DBLP:journals/corr/abs-2006-04152} achieves SOTA results.

We raise two issues for the current SOTA strategy: (1) PABEE faces a limitation for application: it can not flexibly adjust the speedup ratio on a given task and fixed patience parameter, mainly caused by a strict cross-layer comparison strategy. Thus, we wonder whether we can combine PABEE with a softer cross-layer comparison strategy. (2) Current early exiting strategies mainly focus on SLC tasks, while the MLC tasks are neglected. So can they speed up MLC tasks?

Therefore, we propose a Flexible-Patience-Based Early Exiting method (F-PABEE) to address the above issues. F-PABEE makes predictions at each classifier and will exit early if the current layer and the last few layers have similar (similarity score less than a threshold) predicted distributions. F-PABEE can be seen as a natural extension of PABEE and is more flexible since it can achieve better speed-accuracy tradeoffs by adjusting the similarity score thresholds and patience parameters. It can also extend to MLC tasks effortlessly. 

Our contributions are summarized as follows: (1) We propose F-PABEE, a novel and effective inference mechanism that is flexible in adjusting the speedup ratios of PLMs. (2) The results show that our method can accelerate inference effectively while maintaining good performances across different SLC and MLC tasks. (3) We are the first to investigate the early exiting of MLC tasks, and F-PABEE is suitable for this type of task.

\begin{figure*}
    \centering
    \includegraphics[width=12cm,height=6cm]{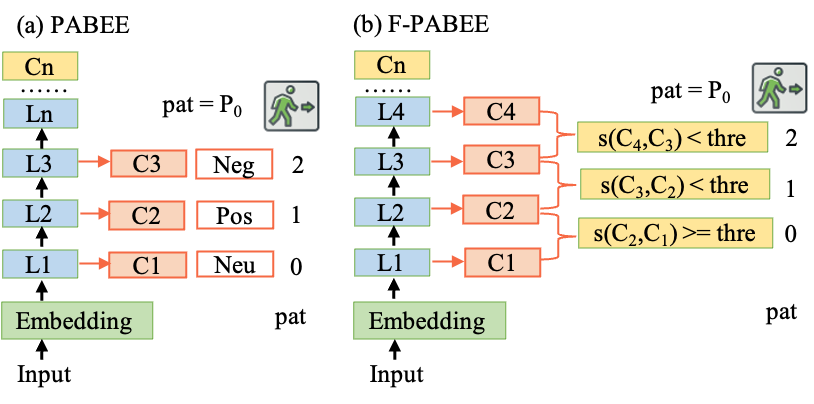}
    \caption{Inference procedure of PABEE and F-PABEE, $C_{i}$ is the classifier, $thre$ is threshold, $P_{0}$ is pre-defined patience.}
    \label{fig1}
\end{figure*}

\section{Related works}
\subsection{Static inference approach}

The static inference approach compresses the heavy model into a smaller one, including pruning, knowledge distillation, quantization, and weight sharing~\cite{Xu2020Bert-of-theseus, DBLP:journals/corr/abs-1910-01108,DBLP:journals/corr/abs-1909-11556}. For example, HeadPrune~\cite{DBLP:journals/corr/abs-1905-10650} ranks the attention heads and prunes them to reduce inference latency. PKD~\cite{DBLP:journals/corr/abs-1908-09355} investigates the best practices of distilling knowledge from BERT into smaller-sized models. I-BERT~\cite{DBLP:journals/corr/abs-2101-01321} performs an end-to-end BERT inference without any floating point calculation. ALBERT~\cite{DBLP:journals/corr/abs-1909-11942} shares the cross-layer parameters. \cite{AutoNLU}, \cite{Zhang2021AutomaticSN} and \cite{Zhu2021DiscoveringBM} distills knowledge from the larger BERT teacher model for improving the performances of student networks which are learned with neural architecture search. Note that the static models are still in the form of deep neural networks with multiple stacked layers. The computational path is invariable for all examples in the inference process, which is not flexible.

\subsection{Dynamic early exiting}

Orthogonal to the static inference approach, early exiting dynamically adjusts hyper-parameters in response to changes in request traffic. It does not need to make significant changes to the original model structure or weight bits, nor does it need to train different teacher-student learning networks, which saves computing resources~\cite{DBLP:journals/corr/abs-1807-03819}.

There are mainly three groups of dynamic early exiting strategies. The first type is confidence-based early exiting~\cite{Zhang2022PCEEBERTAB}. For example, BranchyNet~\cite{7900006}, FastBERT~\cite{DBLP:journals/corr/abs-2004-02178}, and DeeBERT~\cite{DBLP:journals/corr/abs-2004-12993} calculate the entropy of the prediction probability distribution to estimate the confidence of classifiers to enable dynamic early exiting. Shallow-deep~\cite{DBLP:journals/corr/abs-1810-07052} and RightTool~\cite{DBLP:journals/corr/abs-2004-07453} leverage the maximum of the predicted distribution as the exiting signal. The second type is the learned-based exiting, such as BERxiT~\cite{xin-etal-2021-berxit} and CAT~\cite{DBLP:journals/corr/abs-2104-08803}. They learn a criterion for early exiting. The third type is patience-based early exiting, such as PABEE~\cite{DBLP:journals/corr/abs-2006-04152}, which stops inference and exits early if the classifiers' predictions remain unchanged for pre-defined times. Among them, patience-based PABEE achieves SOTA performance. However, PABEE suffers from too strict cross-layer comparison, and the applications on MLC tasks are neglected. There are also literature focusing on improving the training of multi-exit BERT, like LeeBERT \cite{zhu-2021-leebert} and GAML-BERT \cite{zhu-etal-2021-gaml}.

F-PABEE is a more flexible extension to PABEE, which can simultaneously adjust the confidence thresholds and patience parameters to meet different requirements. In addition, it outperforms other existing early exiting strategies on both SLC and MLC tasks.

\subsection{Training of multi-exit backbones}

The literature on early exiting focuses more on the design of early exiting strategies, thus neglect the advances of multi-exit backbones' training methods. LeeBERT~\cite{zhu-2021-leebert} employs an adaptive learning method for training multiple exits. GAML-BERT~\cite{zhu-etal-2021-gaml} enhances the training of multi-exit backbones by a mutual learning approach.

\begin{table*}[tb!]
\centering
\resizebox{0.9\textwidth}{!}{
\begin{tabular}{c|cccccccccccccc}
\hline
  & \multicolumn{2}{c}{CoLA} &  \multicolumn{2}{c}{MNLI} &  \multicolumn{2}{c}{MRPC} & \multicolumn{2}{c}{QNLI} & \multicolumn{2}{c}{QQP} & \multicolumn{2}{c}{RTE} & \multicolumn{2}{c}{SST-2} \\ 


& score  & speedup & score  & speedup & score  & speedup & score  & speedup & score  & speedup & score  & speedup & score  & speedup \\
\hline
BERT base  &  54.2 &  0\%  & 83.1  &  0\%  & 86.8 &  0\%   & 89.8 &  0\%   & 89.2 &  0\%    &  69.1  &  0\%    &  91.3  &  0\%    \\
\hline

Fixed-Exit-3L  & 0.0  &  75\%   & 70.0  & 75\%  & 75.8   & 75\%  & 77.4  &  75\%  & 81.8  &  75\%  &   54.7   &  75\%  &   81.0   &   75\%   \\
Fixed-Exit-6L  & 0.0  &  50\%    &  79.6  &  50\%   &  84.7  &   50\%   &   85.3  &    50\%    &   89.3   &    50\%    &   68.1    &    50\%    &   88.6   &    50\%   \\

\hline
\multirow{2}*{BranchyNet}   &   0.0  &  74\%     &  63.8  &  76\%           &  75.7   &  76\%    &  74.2  &  80\%        &  71.6  &  80\%   &   54.7   &   76\%   &  79.9 &  76\% \\
&  0.0    &    51\%     &   78.3   &   53\%   &  83.0  &  52\%   &   87.1    &   47\%    &   89.3  & 50\%  &  67.4  &  47\%  &  88.3   &   49\%  \\

\hline

\multirow{2}*{Shallow-Deep}   &  0.0  &  75\%     &  64.1  &  77\%           &  75.6   &  76\%    &  74.3  &  78\%        &  71.4  &  79\%   &   54.7   &   76\%   &  79.5 &  77\% \\
&  0.0    &    52\%     &   78.2   &   51\%   &  82.8  &  51\%   &   87.2    &   49\%    &   89.6  & 51\%  &  67.2  &  48\%  &  88.4   &   48\%  \\
\hline

\multirow{2}*{BERxiT}   &  0.0  &  76\%     &  63.5  &  76\%           &  75.6   &  76\%    &  73.3  &  78\%        &  68.2  &  80\%   &   55.3   &   77\%   &  79.5 &  76\%  \\
& 12.3    &    52\%     &   78.4   &   51\%   &  82.9  &  51\%   &   87.0    &   48\%    &   89.1  & 49\%  &  67.3  &  47\%  &  88.3   &   49\% \\
\hline

\multirow{2}*{PABEE}   &  0.0  &  75\%     &  63.9  &  77\%           &  75.8   &  75\%    &  73.6  &  81\%        &  68.6  &  82\%   &   55.8   &   75\%   &  79.9 &  77\% \\
&   0.0    &    50\%     &   78.9   &   52\%   &  83.1  &  53\%   &   87.2    &   46\%    &   89.6  & 49\%  &  67.7  &  46\%  &  88.7   &   48\%  \\
\hline
 
\multirow{2}*{\textbf{F-PABEE}}   &  0.0   & 75\%   &  \textbf{66.9}  &  72\%     &     \textbf{81.5}   &    77\%    &  \textbf{76.2}  &  75\%   &  \textbf{79.6}   &  82\%    &  \textbf{56.0}  &  76\%  &  \textbf{80.5}  & 76\%  \\

&  \textbf{13.6}  &  52\%  &  \textbf{83.9}  &  53\%    &  \textbf{87.3}  &   53\%    &   \textbf{88.6}   &  54\%  &   \textbf{90.8}  &  49\%   &  \textbf{68.1}   &  47\%    &  \textbf{92.3}  &   48\%   \\ 
\hline
 
\hline
\end{tabular}}
\caption{\label{tab:main_results}Experimental results of different early exiting methods with BERT backbone on the GLUE benchmark. }

\end{table*}

\section{Flexible patience-based early exiting}

\subsection{Inference procedure for SLC and MLC tasks}

The inference procedure of F-PABEE is shown in Fig \ref{fig1}(b), which is an improved version of PABEE (Fig \ref{fig1}(a)), where $L_{i}$ is the transformer block of the model, $n$ is the number of transformer layers, $C_{i}$ is the inserted classifier layer, $s$ is the cross-layer similarity score, $thre$ is the similarity score threshold, $P_{0}$ is the pre-defined patience value in the model.

The input sentences are first embedded as the vector:
\begin{equation}
h_0=\text{Embedding}(x). 
\end{equation}
The vector is then passed through transformer layers ($L_{1}...L_{n}$) to extract features and compute its hidden state $h$. After which, we use internal classifiers ($C_{1}...C_{n}$), which are connected to each transformer layer to predict probability $p$:
\begin{equation}
  p_i=C_i (h_i) = C_{i} (L_i (h_{i-1})).
\end{equation}

We denote the similarity score between the prediction results of layer $i-1$ and $i$ as $ s(p_{i-1}, p_{i})$ ($s(p_{i-1}, p_{i}) \in \mathbf{R}$). The smaller the value of  $ s(p_{i-1}, p_{i})$, the prediction distributions are more consistent with each other. The premise of the model's early exit is that the comparison scores between successive layers are relatively small; The similarity threshold $thre$ is a hyper-parameter. We use $pat_i$ to store the times that the cross-layer comparison scores are consecutively less than the threshold $thre$ when the model reaches current layer $i$:

\begin{equation}
pat_{i} = \left\{
\begin{array}{rcl}
 pat_{i-1} + 1&& s(p_{i-1}, p_{i}) < thre\\
0 &&s(p_{i-1}, p_{i}) >= thre 
\end{array}\right\}
\end{equation}

If $ s(p_{i-1}, p_{i})$ is less than the similarity score threshold $thre$, then increase the patience counter by 1. Otherwise, reset the patience counter to 0. This process is repeated until $pat$ reaches the pre-defined patience value $P_0$. The model dynamically stops inference and exits early. However, if this condition is never met, the model uses the final classifier layer to make predictions. This way, the model can stop inference early without going through all layers.

\subsection{Similarity measures for SLC and MLC tasks}

Under the framework of F-PABEE, we can adopt different similarity measures for predicted probability distributions. This work uses the knowledge distillation objectives as the similarity measures~\cite{Hinton2015DistillingTK}. When the model reaches the current layer $l$, for SLC tasks, we compare a series of similarity measures of F-PABEE, denoted as:

\noindent F-PABEE-KD: It adopts the knowledge distillation objective from probability mass distribution $p^{l-1}$ to $p^{l}$:
\begin{equation}
s(p^{l-1},p^{l}) = -\sum_{j=1}^kp_{j}^{l-1}log(p_{j}^{l});
\end{equation}

\noindent F-PABEE-ReKD: It adopts the knowledge distillation objective in the reverse direction, from probability mass distribution $p^{l}$ to $p^{l-1}$:
\begin{equation}
s(p^{l},p^{l-1}) = -\sum_{j=1}^kp_{j}^{l}log(p_{j}^{l-1});
\end{equation}

\noindent F-PABEE-SymKD: It adopts a symmetrical knowledge distillation objective:
\begin{equation}
SymKD = s(p^{l-1},p^{l}) + s(p^{l},p^{l-1});
\end{equation}

\noindent F-PABEE-JSKD: It adopts another symmetrical distillation objective, similar to Jenson-Shannon divergence:
\begin{equation}
JSKD = \frac{1}{2}s(p^{l-1}, \frac{p^{l-1}+p^{l}}{2}) + \frac{1}{2}s(p^{l}, \frac{p^{l-1}+p^{l}}{2})
\end{equation}

In addition, for MLC tasks, we transform them into multiple binary classification problems and sum the similarity scores of all categories, and the formulas are denoted as:

\noindent F-PABEE-KD: 
\begin{equation}
s(p^{l-1},p^{l}) = -\sum_{j=1}^k\sum_{i=1}^2p_{ji}^{l-1}log(p_{ji}^{l});
\end{equation}

\noindent F-PABEE-ReKD:
\begin{equation}
s(p^{l},p^{l-1}) = -\sum_{j=1}^k\sum_{i=1}^2p_{ji}^{l}log(p_{ji}^{l-1});
\end{equation}

The formulations of F-PABEE-SymKD and F-PABEE-JSKD for MLC tasks are similar to those of SLC tasks.

\begin{figure*}
    \centering
    \includegraphics[width=12.9cm,height=9.7cm]{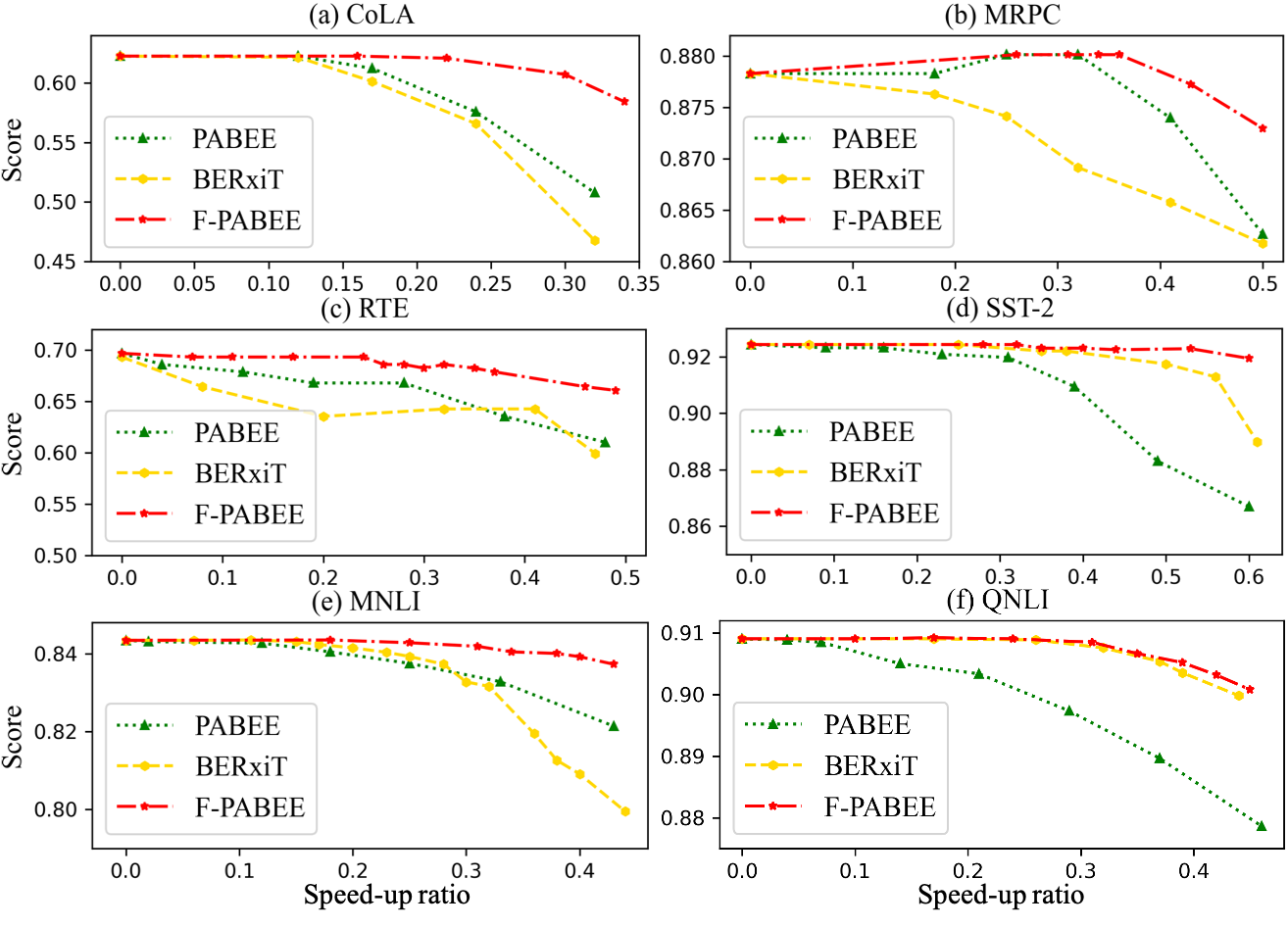}
    \caption{Speed-accuracy curves of F-PABEE, PABEE and BERxiT on SLC tasks with BERT backbone.}
    \label{fig2}
\end{figure*}

\subsection{Training procedure}

F-PABEE is trained on SLC and MLC tasks, while the activation and loss functions are different. For SLC tasks, we use the softmax activation function and cross-entropy function according to the tasks. In contrast, we use the sigmoid activation function and binary cross-entropy function for MLC tasks.

After that, we optimize the model parameters by minimizing the overall loss function $L$, which is the weighted average of the loss terms from all classifiers:

\begin{equation}
L = \sum_{j=1}^njL_{j}/\sum_{j=1}^nj
\end{equation}

\section{Experiments}

\subsection{Tasks and Baselines}

We evaluate F-PABEE on GLUE benchmark~\cite{DBLP:journals/corr/abs-1804-07461} for SLC tasks and four datasets for MLC tasks: MixSNLPS~\cite{DBLP:journals/corr/abs-1805-10190}, MixATS~\cite{hemphill-etal-1990-atis},  AAPD~\cite{DBLP:journals/corr/abs-1806-04822}, and Stackoverflow~\cite{stackoveflow}. we compare F-PABEE with three groups of baselines: (1) BERT-base; (2) Static exiting; (3) Dynamic exiting methods, including BrachcyNet~\cite{Bai2020BinaryBERT}, Shallow-Deep~\cite{DBLP:journals/corr/abs-1810-07052}, BERxiT~\cite{xin-etal-2021-berxit}, and PABEE. Considering the flops of inferencing one with the whole BERT as the base, the speed-up ratio is defined as the average ratio of reduced flops due to early exiting.

\subsection{Experimental setting}

In training process, we perform grid search over the batch size of \{16, 32, 128\}, and learning rate of \{1e-5, 2e-5, 3e-5, 5e-5\} with an AdamW optimizer~\cite{Loshchilov2019DecoupledWD} . The batch size in the inference process is 1. We implement F-PABEE on the bases of HuggingFace Transformers~\cite{wolf2020transformers}. All experiments are conducted on two Nvidia TITAN X 24GB GPUs.

\subsection{Overall comparisons}

\begin{figure*}
    \centering
    \includegraphics[width=12.9cm,height=6cm]{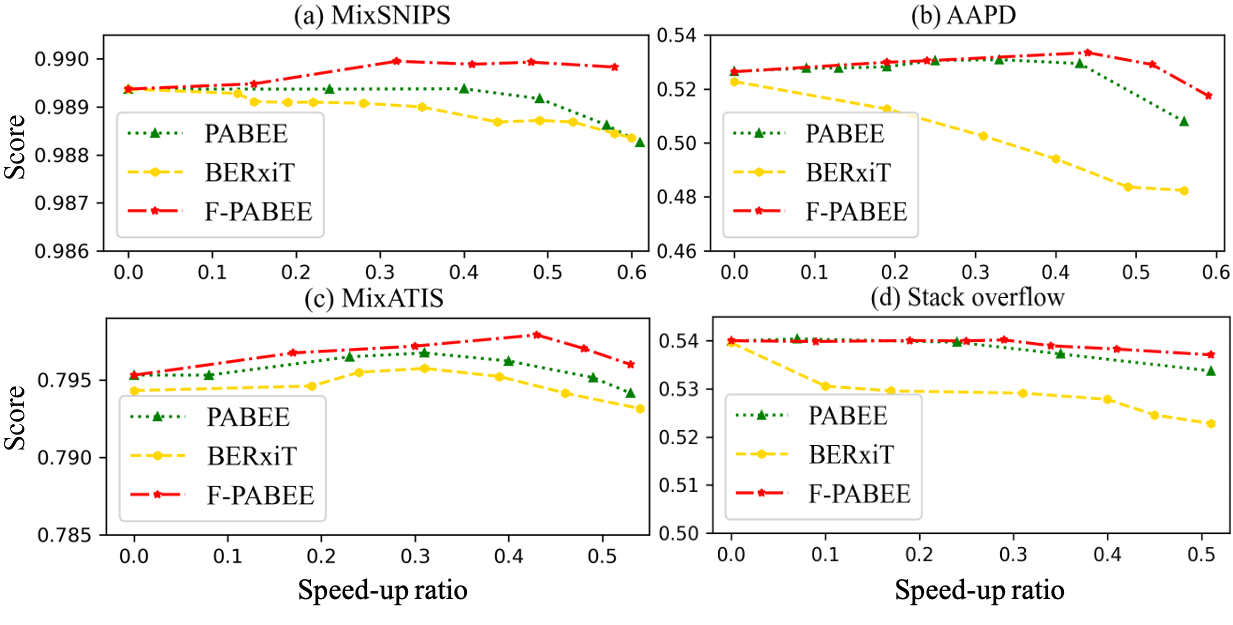}
    \caption{Speed-accuracy curves of F-PABEE, PABEE and BERxiT on MLC tasks with BERT backbone. }
    \label{fig3}
\end{figure*}

\begin{figure*}
    \centering
    \includegraphics[width=12.9cm,height=5.7cm]{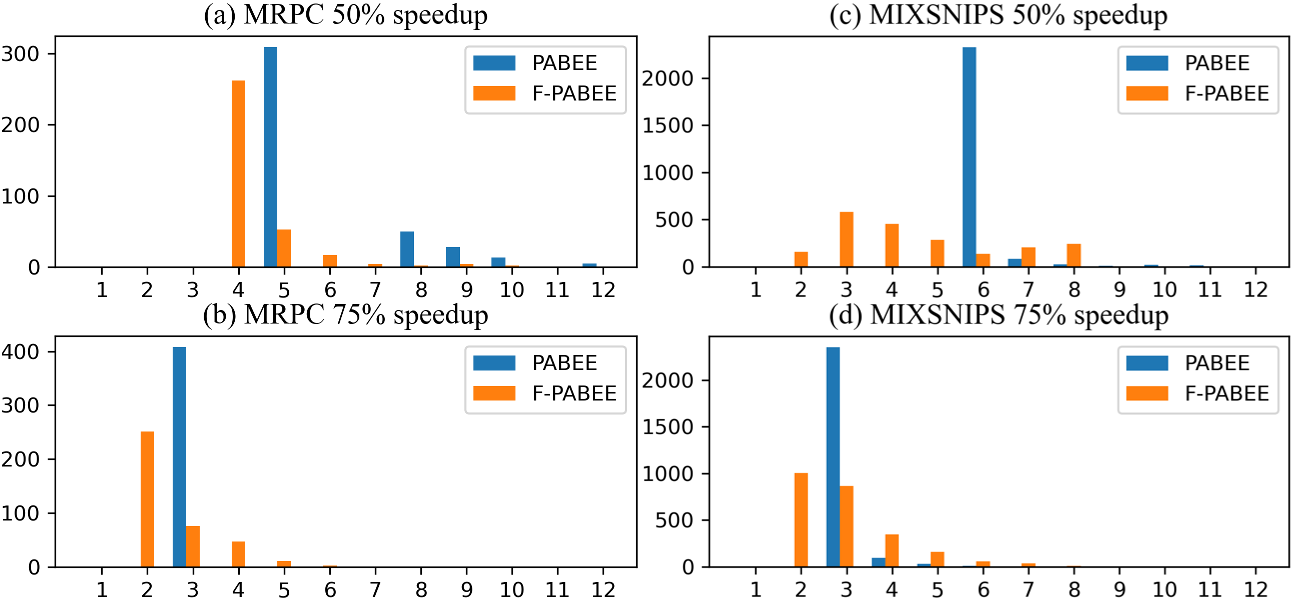}
    \caption{The distribution of executed layers of MRPC and MixSNIPS on average at different speeds (50\%, 75\%). }
    \label{fig6}
\end{figure*}

\begin{figure*}
    \centering
    \includegraphics[width=12.9cm,height=6cm]{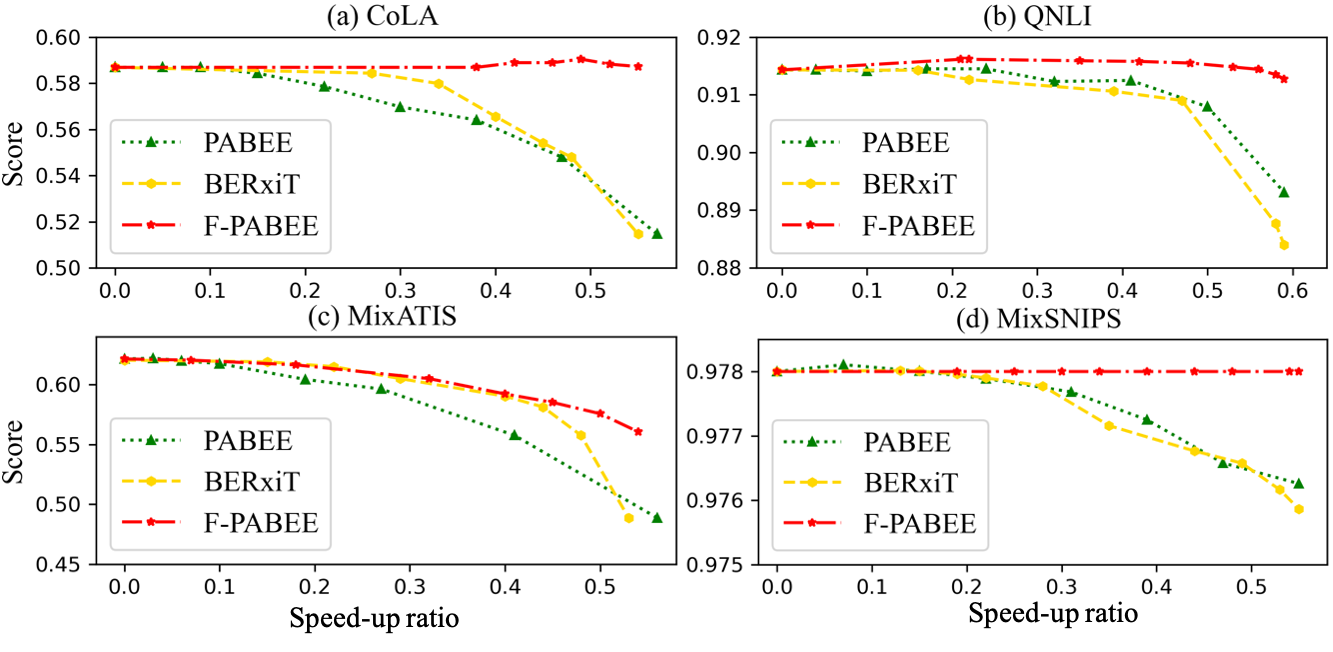}
    \caption{Speed-accuracy curves of F-PABEE, PABEE and BERxiT on SLC and MLC tasks with ALBERT backbone. }
    \label{fig4}
\end{figure*}

\begin{figure*}
    \centering
    \includegraphics[width=12.9cm,height=6cm]{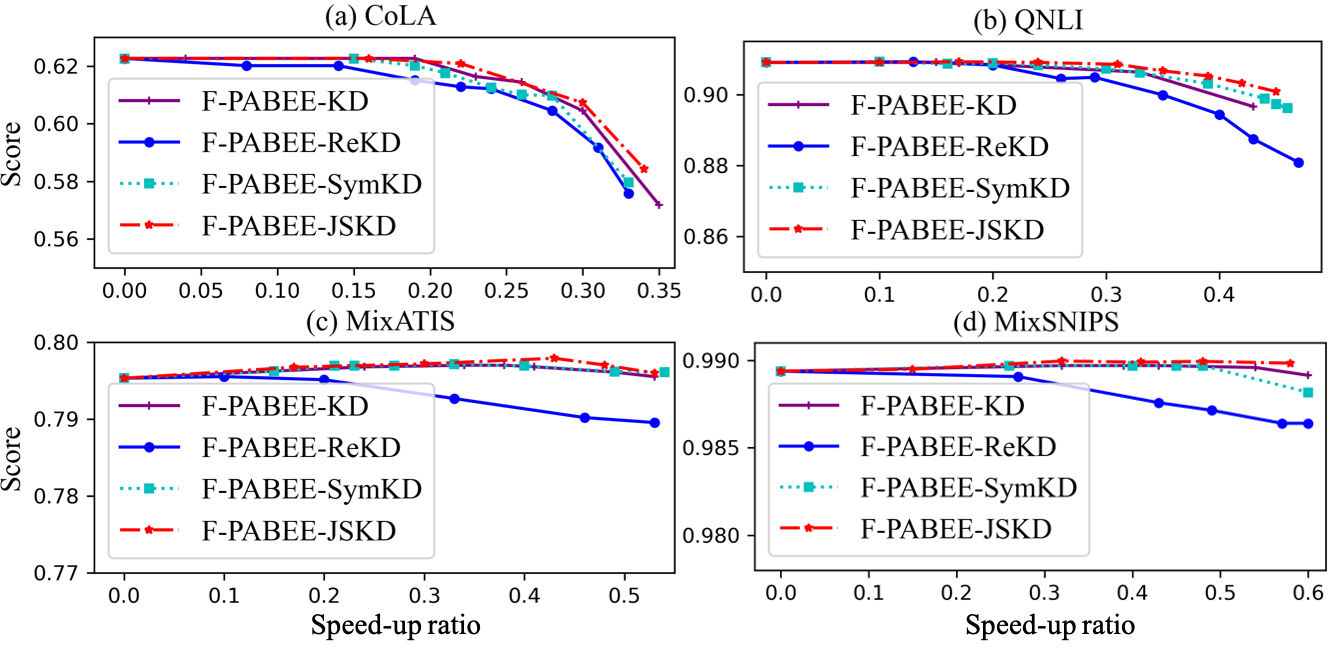}
    \caption{Speed-accuracy curves of different similarity measures on SLC and MLC tasks with BERT backbone. }
    \label{fig5}
\end{figure*}

In Table \ref{tab:main_results}, we compare F-PABEE with other early exiting strategies. We adjust the hyper-parameters of F-PABEE and other baselines to ensure similar speedups with PABEE. It shows that F-PABEE balances speedup and performance better than baselines, especially for a large speedup ratio. Moreover, we draw the score-speedup curves for BERxiT, PABEE, and F-PABEE. It shows that F-PABEE outperforms the baseline models on both SLC (Fig \ref{fig2}) and MLC tasks(Fig \ref{fig3}). Furthermore, the distribution of executed layers (Fig \ref{fig6}) indicates that F-PABEE can choose the faster off-ramp and achieve a better trade-off between accuracy and efficiency by flexibly adjusting similarity score thresholds and patience parameters.

\subsection{Ablation studies}

\textbf{Ablation on different PLMs} \quad F-PABEE is flexible and can work well with other pre-trained models, such as ALBERT. Therefore, to show the acceleration ability of F-PABEE with different backbones, we compare F-PABEE to other early exiting strategies with ALBERT base as the backbone. The results in Fig \ref{fig4} show that F-PABEE outperforms other early exiting strategies under different backbones by large margins on both SLC and MLC tasks, indicating that F-PABEE can accelerate the inference process for numerous PLMs.

\textbf{Comparisons between different similarity measures} \quad We consider F-PABEE with different similarity measures, denoted as F-PABEE-KD, F-PABEE-ReKD, F-PABEE-SymKD, and F-PABEE-JSKD, and the results are presented in Fig \ref{fig5}. F-PABEE-JSKD performs the best on both SLC and MLC tasks. We suppose that F-PABEE-JSKD is symmetric, and the similarity discrimination is more accurate than asymmetric measures. Therefore, it is better at determining which samples should exit at shallow layers and which should go through deeper layers.

\section{Conclusions}

We proposed F-PABEE, a novel and efficient early exiting method that combines PABEE with a softer cross-layer comparison strategy. F-PABEE is more flexible than PABEE since it can achieve different speed-performance tradeoffs by adjusting the similarity score thresholds and patience parameters. In addition, we investigate the acceleration ability of F-PABEE with different backbones. Moreover, we compare the performances of F-PABEE with different similarity measures. Extensive experiments on SLC and MLC demonstrate that: (1) F-PABEE performs better than the previous SOTA adaptive early exiting strategies for both SLC and MLC tasks. As far as we know, we are the first to investigate the early exiting methods for MLC tasks. (2) F-PABEE performs well on different PLMs such as BERT and ALBERT. (3) Ablation studies show that F-PABEE-JSKD performs best for F-PABEE with different similarity measures.

\vfill\pagebreak

\bibliographystyle{IEEEbib}
\bibliography{strings,refs}

\end{document}